\documentclass[runningheads]{llncs}
\usepackage{multirow}
\usepackage{graphicx}
\usepackage{amsmath,amssymb} 
\usepackage{color}

\begin{document}
%
\title{Dense Pose Transfer}

\titlerunning{Dense Pose Transfer}
%
\author{Natalia Neverova\inst{1} \and 
R{\i}za Alp G\"uler\inst{2} \and 
Iasonas Kokkinos\inst{1}}
%
\authorrunning{N. Neverova, R. A. G\"uler, I. Kokkinos}
%

\institute{Facebook AI Research, Paris, France,
\email{\{nneverova, iasonask\}@fb.com} \and
INRIA-CentraleSup\'elec, Paris, France, 
\email{riza.guler@inria.fr}}
\maketitle              
%
\begin{abstract}
In this work we integrate ideas from surface-based modeling with neural synthesis: we propose a combination of surface-based pose estimation and deep generative models
that allows us to perform accurate pose transfer, i.e. synthesize a new image of a person based on a single image of that person and the image of a pose donor.
We use a dense pose estimation system that maps pixels from both images to a common surface-based coordinate system, allowing the 
two images to be  brought in correspondence with each other.  
We inpaint and refine the source image intensities in the surface coordinate system, prior to warping them onto the target pose. These predictions are fused with those of a convolutional predictive module through a neural synthesis module allowing for training the whole pipeline jointly end-to-end, optimizing a combination of adversarial and perceptual losses. 
We show that dense pose estimation is a substantially more powerful conditioning input than landmark-, or  mask-based alternatives, and report systematic improvements over state of the art generators on DeepFashion and MVC datasets. 
\end{abstract}
\begin{figure}[!h]
\centering
\includegraphics[width=0.95\linewidth]{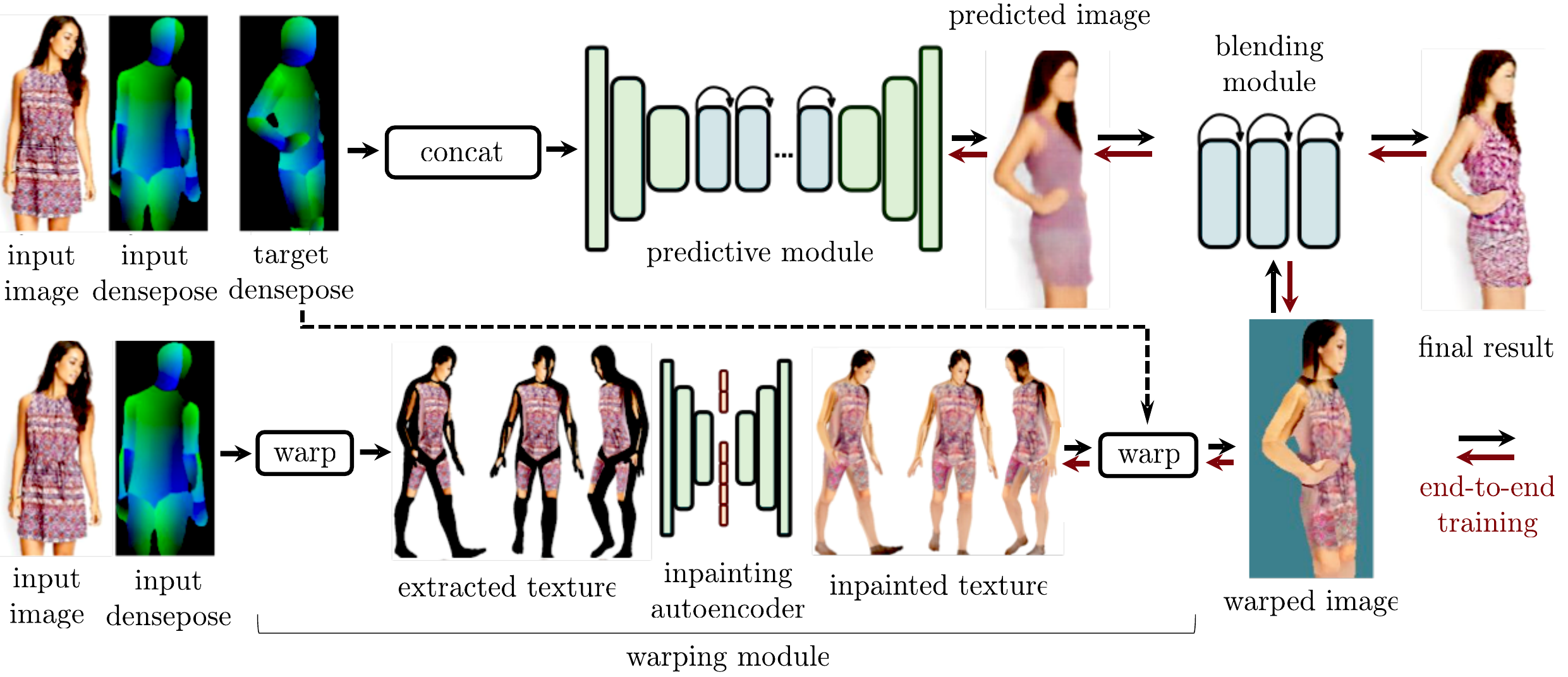} 
\caption{Overview of our pose transfer pipeline: 
given an input image and a target pose we use DensePose \cite{DensePose} to drive the generation process. This is achieved through the complementary streams of (a) a data-driven predictive module, and (b) a surface-based module that warps the texture to UV-coordinates, interpolates on the surface, and warps back to the target image. A blending module combines the complementary merits of these two streams in a single end-to-end trainable framework.} 
\label{fig:pipeline}
\end{figure}

\section{Introduction}
\label{sec:Introduction}
Deep models have recently shown remarkable success in tasks such as face \cite{KarrasAilaLaine2017}, human \cite{LassnerPonsMollGehler2017,PoseGuided,SiarohinSangineto2017}, or scene generation  \cite{ChenKoltun2017,WangLiuZhu2017}, collectively known as ``neural synthesis''. 
This opens countless possibilities in computer graphics applications, including cinematography, gaming and virtual reality settings. At the same time, the potential malevolent use of this technology raises new research problems, including the detection of forged images or videos \cite{forensics2018}, which in turn requires training forgery detection algorithms with multiple realistic samples.
In addition, synthetically generated images have been successfully exploited for data augmentation and training deep learning frameworks for relevant recognition tasks~\cite{ShrivastavaPfisterCVPR2017}.


In most applications, the relevance of a generative model to the task directly relates to the amount of  control that one can exert on the generation process.
Recent works have shown the possibility of adjusting image synthesis by controlling categorical attributes \cite{fader,LassnerPonsMollGehler2017}, low-dimensional parameters \cite{NeuralFace2017}, or layout constraints indicated by a conditioning input \cite{pix2pix,ChenKoltun2017,WangLiuZhu2017,LassnerPonsMollGehler2017,PoseGuided,SiarohinSangineto2017}. 
In this work we aspire to obtain a stronger hold of the image synthesis process by relying on  surface-based object representations, similar to the ones used in graphics engines. 

Our work is focused on the human body, where surface-based image understanding has been most recently unlocked \cite{SMPL:2015,bogo2016keep,Lassner:UP:2017,varol17,kanazawa2017end,DenseReg,DensePose}. 
We build on the recently introduced SMPL model~\cite{SMPL:2015} and the DensePose system \cite{DensePose}, which taken together allow us to interpret an image of a person in terms of a full-fledged surface model, getting us closer to the goal of performing \textit{inverse graphics}. 

In this work we close the loop and perform image generation by rendering the same person in a new pose through surface-based neural synthesis. 
The target pose is indicated by the image of a `pose donor', i.e. another person that guides the image synthesis. The DensePose system is used to associate the new photo with the common surface coordinates and copy the appearance predicted there.  

The purely geometry-based synthesis process is on its own insufficient for realistic image generation: its performance can be compromised by inaccuracies of the DensePose system as well as by self-occlusions  of the body surface in at least one of the two images. We account for occlusions by introducing an inpainting network that operates in the surface coordinate system and combine its predictions with the outputs of a more traditional feedforward conditional synthesis module. These predictions are obtained independently and compounded by a refinement module that is trained so as to optimize  a combination of reconstruction, perceptual and adversarial losses. 


We experiment on the DeepFashion~\cite{deepfashion} and MVC~\cite{MVC} datasets and show that we can obtain results that are quantitatively  better than the latest state-of-the-art. Apart from the specific problem of pose transfer, the proposed combination of neural synthesis with surface-based representations can also be promising for the broader problems of virtual and augmented reality:  the  generation process is more transparent and easy to connect with the physical world, thanks to the underlying surface-based representation. In the more immediate future, the task of pose transfer can be useful for dataset augmentation, training forgery detectors, 
as well as texture transfer applications like those showcased by~\cite{DensePose}, without however requiring the acquisition of a surface-level texture map.

\newcommand{\mycomment}[1]{}
\mycomment{
\begin{equation*}
A_{S}|_{\denseposeusourcevs}  = I_S(\denseposesourcexy)\\
I_T(\denseposexy) = A_{S}(
A_{S}|_{\denseposeuv}  = )\\
\end{equation*}
}

\section{Previous works}
\textbf{Deep generative models} have originally been studied as a means of unsupervised feature learning \cite{hinton2006reducing,kingma2013auto,goodfellow2014generative,radford2015unsupervised}; however,  based on the increasing realism of neural synthesis models \cite{GatysEckerBethgeICLR2015,KarrasAilaLaine2017,ChenKoltun2017,WangLiuZhu2017} such models can now be considered as  components in computer graphics applications such as content manipulation \cite{WangLiuZhu2017}. \smallskip

\textbf{Loss functions} used to train such networks largely determine the realism of the resulting outputs. Standard reconstruction losses, such as $\ell_1$ or $\ell_2$  norms typically result in blurry results, but at the same time enhance stability \cite{pix2pix}. Realism can be enforced by using an 
adapted discriminator loss trained in tandem with a generator in Generative Adversarial Network (GAN) architectures~\cite{goodfellow2014generative} to ensure that the generated and observed samples are indistinguishable. However, this training can often be unstable, calling for more robust variants such as the squared loss of~\cite{LSGAN}, WGAN and its variants~\cite{WGAN} or multi-scale discriminators as in~\cite{WangLiuZhu2017}. An alternative solution is the perceptual loss used in \cite{JohnsonAlahiLiECCV16,UlyanovLebedevICML2016} replacing the optimization-based style transfer of~\cite{GatysEckerBethgeICLR2015}  with feedforward processing. This was recently shown in~\cite{ChenKoltun2017} to deliver substantially more accurate scene synthesis results than~\cite{pix2pix}, while compelling results were obtained more recently by combining this loss with a GAN-style discriminator~\cite{WangLiuZhu2017}. \smallskip 
%
%
\mycomment{
	https://papers.nips.cc/paper/6644-pose-guided-person-image-generation.pdf \cite{PoseGuided}\\
	https://arxiv.org/pdf/1704.04886.pdf \cite{BoWuCheng2017} - variational GANs\\
	https://arxiv.org/pdf/1801.00055.pdf \cite{SiarohinSangineto2017} - deformable GANs (looks great)\\
	segmentations and keypoints \cite{LassnerPonsMollGehler2017}\\
	NVIDIA progressive GANs \cite{KarrasAilaLaine2017}
}

\textbf{Person and clothing synthesis} has  been addressed in a growing body of recently works \cite{LassnerPonsMollGehler2017,SiarohinSangineto2017,PoseGuided,zhu2017your}. All of these works assist the image generation task through domain, person-specific  knowledge, which gives both better quality results, and a more controllable image generation pipeline.\smallskip

\textbf{Conditional neural synthesis} of humans has been shown in 
\cite{SiarohinSangineto2017,PoseGuided}  to provide a strong handle on the output of the generative process. 
A controllable surface-based model of the human body is used in \cite{LassnerPonsMollGehler2017} to drive the generation of  persons wearing clothes with controllable color combinations. The  generated images are demonstrably realistic, but the pose is determined by controlling a surface based model, which can be limiting if one wants e.g. to render a source human based on a target video.  A different approach is taken in the pose transfer work of~\cite{PoseGuided}, where a sparse set of landmarks detected in the target image are used  as a conditioning input to a generative model. The authors show that pose can be generated with increased accuracy, but often losing texture properties of the source images, such as cloth color or texture properties. In the work of \cite{ZhaoWCLF17} multi-view supervision is used to train a two-stage system that can generate images from multiple views. 
In more recent work \cite{SiarohinSangineto2017} the authors show that introducing a correspondence component in a GAN architecture allows for substantially more accurate pose transfer. \smallskip
%

\textbf{Image inpainting} helps estimate the body appearance on occluded body regions. Generative models are able to fill-in information that is labelled as occluded, either by accounting for the occlusion pattern during training
\cite{PathakKDDE16}, or by optimizing a score function that indicates the quality of an image, such as the negative of a GAN discriminator loss \cite{YehCLHD16}.
The work of \cite{YehCLHD16} inpaints arbitrarily occluded faces by  minimizing the discriminator loss of a GAN trained with fully-observed face patches. In the realm of face analysis impressive results have been generated recently by works that operate in the UV coordinate system of the face surface, aiming at photorealistic face inpainting \cite{SaitoWHNL16}, and  pose-invariant identification \cite{uvgan}. Even though we address a similar problem, the lack of access to full UV recordings (as in~\cite{uvgan,SaitoWHNL16}) poses an additional challenge.




\mycomment{
\section{Prior work}
Relevant papers (state-of-the-art):\\
https://papers.nips.cc/paper/6644-pose-guided-person-image-generation.pdf \cite{PoseGuided}\\
https://arxiv.org/pdf/1704.04886.pdf \cite{BoWuCheng2017} - variational GANs\\
https://arxiv.org/pdf/1801.00055.pdf \cite{SiarohinSangineto2017} - deformable GANs (looks great)\\
segmentations and keypoints \cite{LassnerPonsMollGehler2017}\\
NVIDIA progressive GANs \cite{KarrasAilaLaine2017}
}

	\section{Dense Pose Transfer}
	\newcommand{\warping}{warping~}
	\newcommand{\reffig}[1]{Fig.~\ref{#1}}
	\newcommand{\iae}{Inpainting Auto-Encoder}
	\newcommand{\refsec}[1]{Sec.~\ref{#1}}
	
	
	We develop our approach to pose transfer around the \textit{DensePose} estimation system~\cite{DensePose} to associate  every human pixel with its coordinates on a surface-based parameterization of the human body in an efficient, bottom-up manner.
	We exploit the DensePose outputs in two complementary ways, corresponding to the {\em predictive module} and
	the {\em \warping  module}, as shown in~\reffig{fig:pipeline}.
	The warping module uses DensePose surface correspondence and inpainting to generate a new view of the person, while the predictive module is a generic, \textit{black-box}, generative model conditioned on the DensePose outputs for the input and the target. 
	
	These modules corresponding to two parallel streams have complementary merits: the \textit{predictive module} successfully exploits the dense conditioning output to generate plausible images for familiar poses, delivering superior results to those obtained from sparse, landmark-based conditioning; at the same time, it cannot generalize to new poses, or transfer texture details. 
	By contrast,  the \textit{warping module} can preserve  high-quality details and textures, allows us to perform inpainting in a uniform, canonical coordinate system, and generalizes for free for a broad variety of body movements. However, its body-, rather than clothing-centered construction does not take into account hair, hanging clothes, and accessories. 
	The best of both worlds is obtained by feeding the outputs of these two blocks into a {\em blending module} trained to fuse and refine their predictions using a combination of reconstruction, adversarial, and perceptual losses in an end-to-end trainable framework. 
	
	
	The DensePose module is common to both streams and delivers dense correspondences between an image and a surface-based model of the human body. It does so by firstly  assigning every pixel to one of 24 predetermined surface parts,  and then regressing the part-specific surface coordinates of every pixel.
	The results of this system are encoded in three output channels, comprising the part label and part-specific UV surface coordinates. 
	 This system is trained discriminatively and provides a simple, feed-forward module for dense correspondence from an image to the human body surface.
	We omit further details, since we rely  on the system of \cite{DensePose} with minor implementation differences   described in Sec.~\ref{experiments}.\medskip
	
	Having outlined the overall architecture of our system, in \refsec{predictive} and \refsec{surface}  we present  in some more detail our components, and then turn in \refsec{losses} to the loss functions used in their training. A thorough description of  architecture details is left to the supplemental material. 
	We start by presenting the architecture of the predictive stream, and then turn to the surface-based stream, corresponding to the upper and lower rows of \reffig{fig:pipeline}, respectively.
%
%
	\subsection{Predictive stream}
	\label{predictive}
	
	The predictive module is a conditional generative model that exploits the DensePose system results for pose transfer. Existing conditional models indicate the target pose in the form of  heat-maps from keypoint detectors~\cite{PoseGuided}, or part segmentations~\cite{LassnerPonsMollGehler2017}.  Here we condition on the concatenation of  the input image and  DensePose results for the input and  target images,  resulting in an input of  dimension $256{\times}256{\times}9$. This  provides  conditioning  that is both global (part-classification), and point-level (continuous coordinates), allowing the remaining network to exploit a richer source of information. 
	
	The remaining architecture includes an encoder followed by a stack of residual blocks and a decoder at the end, along the lines of  \cite{JohnsonAlahiLiECCV16}. 
	In more detail, this network comprises (a) a cascade  of three  convolutional layers that  encode the  $256{\times}256{\times}9$ input into $64{\times}64{\times} 256$  activations,  (b) a set of six residual blocks with $3{\times}3{\times}256{\times}256 $ kernels, (c) a cascade of two deconvolutional and one convolutional layer that deliver an output of the same spatial resolution as the input. 
	All intermediate convolutional layers have $3{\times}3$ filters and are followed by instance normalization \cite{instanceNorm} and ReLU activation. The last layer has tanh non-linearity and no normalization.
	%
	%
%
	\smallskip
	\subsection{Warping stream}
	\label{surface}
\begin{figure}[t!]
\centering
\includegraphics[width=0.88\linewidth]{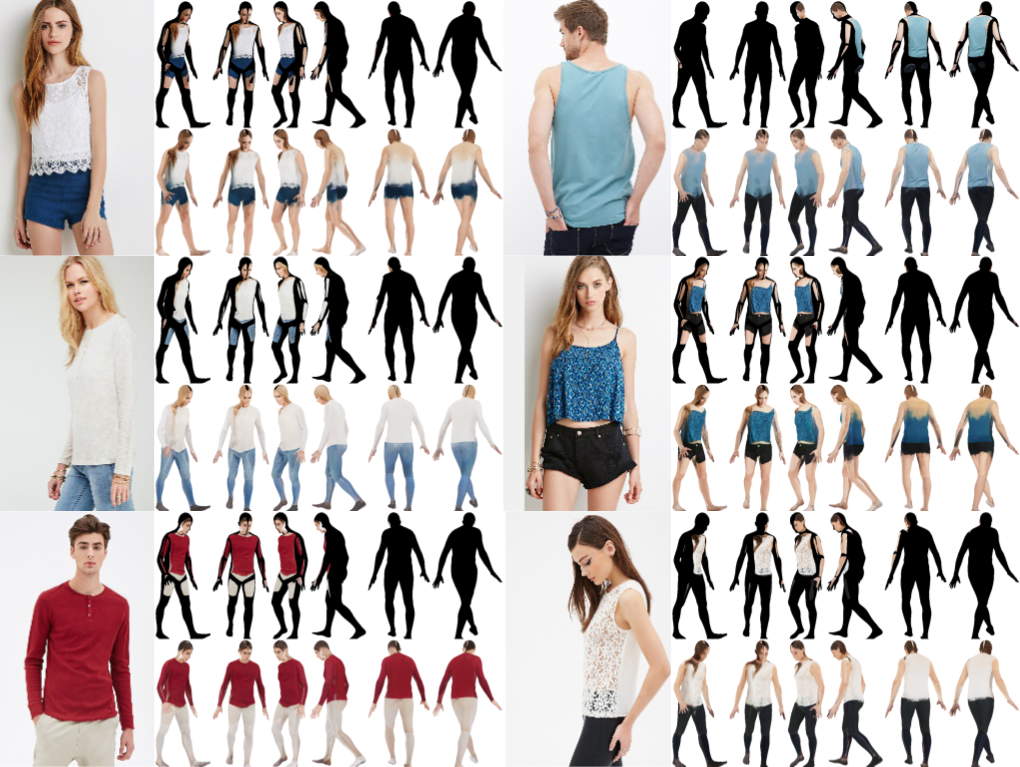}
\caption{Warping module results. 
For each sample, the top row shows interpolated textures obtained from DensePose predictions and projected on the surface of the 3D body model. The bottom row shows the same textures after inpainting in the UV space.}
\label{fig:warpingresults}
\end{figure}

	Our warping module  performs pose transfer by performing explicit texture mapping between the input and the target image on the common surface UV-system. 
	\mycomment{
		\begin{equation*}
		A_{S}|_{\denseposeusourcevs}  = I_S(\denseposesourcexy)\\
		I_T(\denseposexy) = A_{S}(
		A_{S}|_{\denseposeuv}  = )\\
		\end{equation*}
	}
	The core of this component is a Spatial Transformer Network (STN)~\cite{STN}  that warps according to   DensePose  the  image observations  to the  UV-coordinate system of each surface part; we use a  grid with $256 {\times} 256 $ UV points for each of the 24 surface parts, and perform scattered interpolation to 	handle the continuous values of the regressed UV coordinates. The inverse mapping from UV to the output image space is performed by a second STN with a bilinear kernel.
	%
	%
	As shown in \reffig{fig:warpingresults}, a direct implementation of this module would often deliver poor results:
	the part of the surface that is visible on the source image is typically small, and can often be entirely non-overlapping with the part of the body that is visible on the target image. This is only exacerbated by DensePose failures or systematic errors around the part seams.
	These problems motivate the use of an inpainting  network within the warping module, as detailed below.\smallskip
\begin{figure}[!h]
\centering
\includegraphics[width=\linewidth]{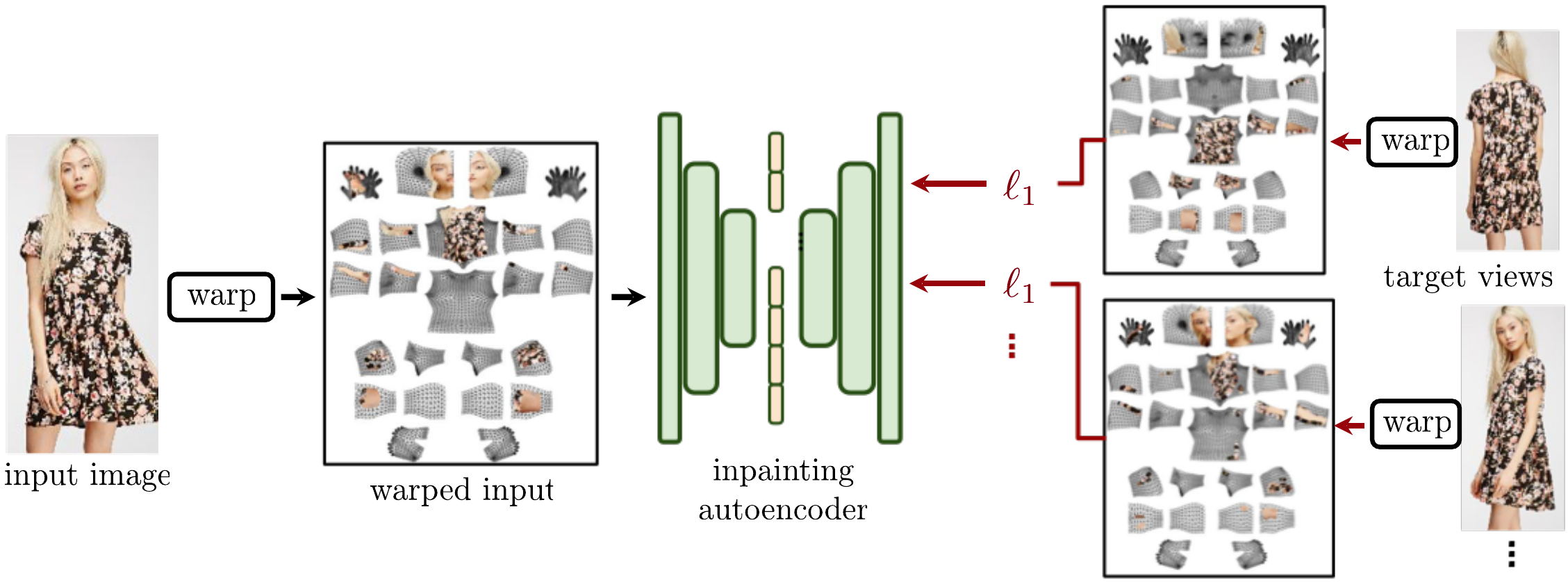} 
\caption{Supervision signals for pose transfer on the warping stream: the input image on the left is warped to intrinsic surface coordinates through a spatial transformer network driven by DensePose. From this input, the inpainting autoencoder has to predict the appearance of the same person from different viewpoints, when also warped to intrinsic coordinates. The loss functions on the right penalize the reconstruction only on the observed parts of the texture map. This form of multi-view supervision acts like a surrogate for the (unavailable) appearance of the person on the full body surface. 
}
\label{fig:inpainting}
\end{figure}
	
	\textbf{Inpainting autoencoder.}
	This model allows us to extrapolate the body appearance from the  surface nodes populated by the STN to the remainder of the surface. 
	Our setup requires a different approach to the one of other deep inpainting methods \cite{YehCLHD16},
	because we never observe the full surface texture during training.
We handle the partially-observed nature of our training signal by using a reconstruction  loss that only penalizes the observed part of the UV map, and lets the network freely \textit{guess} the remaining domain of the signal. In particular, we use a masked $\ell_1$ loss on the difference between the autoencoder predictions and the target signals, where the masks indicate the visibility of the target signal. 

We observed that by its own this does not urge the network to inpaint successfully;  results substantially improve when we accompany every input with multiple supervision signals, as shown in \reffig{fig:inpainting}, corresponding to UV-wrapped shots of the same person at different poses. This fills up a larger portion of the UV-space and forces the inpainting network to predict over the whole texture domain. 
	As shown in \reffig{fig:warpingresults}, the inpainting process allows us to obtain a uniformly observed surface, which captures the  appearance  of skin and tight clothes, but does not account for hair, skirts, or apparel, since these are not accommodated by  DensePose's surface model.

Our inpainting network is comprised of $N$ autoencoders, corresponding to the decomposition of the body surface into $N$ parts used in the original DensePose system \cite{DensePose}. This is based on the observation that appearance properties are non-stationary over the body surface. 
Propagation of the context-based information from visible to invisible parts that are entirely occluded at the present view is achieved through a fusion mechanism that operates  at the level of latent representations delivered by the individual encoders, and injects global pose context in the individual encoding through a concatenation operation.

In particular, we denote by $\mathcal{E}_i$ the individual encoding delivered by the encoder for the $i$-th part. 
The fusion layer 
concatenates these obtained encodings into a single vector  
which is then down-projected to a 256-dimensional global pose embedding through a linear layer.
We  pass the resulting embedding through a cascade of ReLU and Instance-Norm units and  transform it again into an embedding denoted by $\mathcal{G}$. 

Then the $i$-th part decoder receives as an input the concatenation of~$\mathcal{G}$ with~$\mathcal{E}_i$,
which  combines information particular to part $i$, and global context, delivered by $\mathcal{G}$. This is processed by a stack of 
deconvolution operations, which delivers in the end the prediction for the texture of part $i$.

%
%
	%
	%
	%
	\subsection{Blending module}
	\label{surface}
	The blending module's objective is to combine the complementary strengths of the two streams to deliver a single fused result, that will be `polished' as measured by the losses used for training. As such it no longer involves an encoder or decoder unit, but rather only contains two convolutional and three residual blocks that aim at combining the predictions and refining their results. 
	
	In our framework, both predictive and warping modules are first pretrained separately and then finetuned jointly during blending.
	The final refined output is obtained by learning a residual term added to the output of the predictive stream. The blending module takes an input consisting of the outputs of the predictive and the warping modules combined with the target dense pose. 
	
	
\subsection{Loss Functions}
\label{losses}
\newcommand{\esty}{\boldsymbol{\hat{y}}}
\newcommand{\by}{\boldsymbol{y}}
\newcommand{\bz}{\boldsymbol{z}}

As shown in \reffig{fig:pipeline}, the training set for our network comes in the form of pairs of input and target images,  $\boldsymbol{x}$, $\boldsymbol{y}$ respectively, both of which  are of the same person-clothing, but in different poses.  Denoting by $\boldsymbol{\hat{y}}=G(\boldsymbol{x})$ the network's prediction, the difference between  $\boldsymbol{\hat{y}},\boldsymbol{y}$ can be measured through a multitude of loss terms,  that penalize different forms of deviation. We present them below for completeness 
and ablate their impact in practice in Sec.~\ref{experiments}.
\smallskip

	\textbf{Reconstruction loss.} To penalize reconstruction errors we use the common~$\ell_1$  distance between the two signals: $\|\boldsymbol{\hat{y}} -\boldsymbol{y}\|_1$. On its own, this delivers blurry results but is important for retaining the overall intensity levels.  
\smallskip	

	\textbf{Perceptual loss.} As in Chen and Koltun~\cite{ChenKoltun2017}, we use a  VGG19 network pretrained for classification \cite{VGG} 
as a feature extractor for  both	$\boldsymbol{\hat{y}},\boldsymbol{y}$ and penalize the~$\ell_2$ distance of the respective intermediate feature activations $\Phi^{v}$ at 5 different network layers $v=1,\ldots,N$:
	\begin{equation}
\mathcal{L}_{\mathrm{p}}(\boldsymbol{y},\boldsymbol{\hat{y}}) = \sum_{v=1}^{N}\|{\Phi}^{v}(\boldsymbol{y}) - {\Phi}^{v}(\boldsymbol{\hat{y}})\|_2.
\label{perceptual}
\end{equation}
This loss penalizes differences in  low- mid- and high-level feature statistics, captured by the respective network filters. 
\smallskip

\textbf{Style loss.} As in \cite{JohnsonAlahiLiECCV16}, we use the Gram matrix criterion of  \cite{GatysEckerBethgeICLR2015} as an objective for training a feedforward network. 
We first compute the Gram matrix of neuron activations delivered by the VGG network $\Phi$ at layer $v$ for an image $\boldsymbol{x}$:
\begin{equation}
\mathcal{G}^{v}(\boldsymbol{x})_{c,c'} = 
\sum_{h,w}\Phi^{v}_{c}(\boldsymbol{x})[h,w]\Phi^{v}_{c'}(\boldsymbol{x})[h,w]
\end{equation}
where $h$ and $w$ are horizontal and vertical pixel coordinates and $c$ and $c'$ are feature maps of layer $v$. 
The style loss is given by the sum of  Frobenius norms for the difference between the per-layer Gram matrices $\mathcal{G}^{v}$ of the two inputs:
\begin{equation}
\mathcal{L}_{\mathrm{style}}(\boldsymbol{y},\boldsymbol{\hat{y}}) = \sum_{v=1}^B \|\mathcal{G}^{v}(\boldsymbol{y}) - \mathcal{G}^{v}(\boldsymbol{\hat{y}})\|_{F}.
\end{equation}

\textbf{Adversarial loss.} We use adversarial training to penalize any detectable differences between the generated and real samples. Since global structural properties are largely settled thanks to  DensePose conditioning, we opt for the patchGAN~\cite{pix2pix} discriminator, which operates locally and picks up differences between texture patterns.
The discriminator~\cite{pix2pix,WangLiuZhu2017} takes as an input~$\bz$, a combination of the source image and the DensePose results on the target image, and either the target image $\by$ (real) or the generated output (fake) $\esty$. We want  fake samples to be indistinguishable from real ones -- as such we optimize the following objective:
\begin{equation}
L_{\text{GAN}} = \underbrace{\frac{1}{2}\mathbb{E}_{\bz}\left[l(D(\bz,\by)-1)\right] + \frac{1}{2}\mathbb{E}_{\bz}\left[l(D(\bz,\esty))\right]}_{\mathrm{Discriminator}}
+ \underbrace{\frac{1}{2}\mathbb{E}_{\boldsymbol{z}}\left[l(D(G(\bz)-1))\right]}_{\mathrm{Generator}},
\end{equation}
where we use $l(x) = x^2$ as in the  Least Squares GAN (LSGAN) work of \cite{LSGAN}. 


\section{Experiments}
\label{experiments}

We perform our experiments on the DeepFashion dataset (In-shop Clothes Retrieval Benchmark) \cite{deepfashion} that contains 52,712 images of fashion models demonstrating 13,029 clothing items in different poses. All images are provided at a  resolution of $256{\times}256$ and contain people captured over a uniform background. Following~\cite{SiarohinSangineto2017} we select 12,029 clothes for training and the remaining 1,000 for testing. 
For the sake of direct comparison with state-of-the-art keypoint-based methods, we also remove all images where the keypoint detector of \cite{CMUdetector} does not detect any body joints. This results in 140,110 training and 8,670 test pairs.

In the supplementary material we provide results on the large scale MVC dataset~\cite{MVC} that consists of 161,260 images of resolution $1920{\times}2240$ crawled from several online shopping websites and showing front, back, left, and right views for each clothing item.

\subsection{Implementation details}
    %
    %
    \textbf{DensePose estimator}.
    We use a fully convolutional network (FCN) similar to the one used as a teacher network in \cite{DensePose}. The FCN is a ResNet-101 trained on cropped person instances from the COCO-DensePose dataset. 
    %
    The output 
    consists of 2D fields 
    representing body segments (I) and $\{U,V\}$ coordinates in coordinate spaces aligned with each of the semantic parts of the 3D model. 
    %
    %
    \mycomment{
    \item[] \textb {Predictive module}.
    The architecture of this module is similar to the image transformation network by Johnson et al. \cite{JohnsonAlahiLiECCV16} and includes three functional blocks: 1) a set of convolutional layers downsampling the input by a factor of 4, 2) a set of 6 residual blocks with 256 filters, 3) an inverted input block recovering the original dimensionality of the input in the output.
    All convolutional layers have $3{\times}3$ filters and are followed by instance normalization \cite{instanceNorm} and ReLU activation (except the last layer, that has tanh non-linearity and no normalization).
    Denoting $CN_{m}^n$ ($DCN_{m}^n$) a convolutional (deconvolutional) block with activation and normalization, m filters and stride n, the architectures of input and output blocks can be written as $CN_{64}^1$-$CN_{128}^2$-$CN_{256}^2$ and $DCN_{128}^2$-$DCN_{64}^2$-$CN_{3}^1$, respectively.
    The predictive module takes as an input a concatenation of the input image and DensePose representations of the input and the target poses (resulting dimension $256{\times}256{\times}9$) and outputs an image of size $256{\times}256{\times}3$.
    \bigskip\\
    \textbf{Warping module}.
    The structure of the warping module is shown in Fig.~\ref{fig:warping}. Using a Spatial Transformer Network (STN)~\cite{STN} followed by scattered interpolation, the source image is first projected onto UV (texture) space of dimensionality $N_u{\times}N_v{\times}N$, where $N_u{\times}N_v$ is texture resolution and $N$ is the number of body segments in the model ($N_u{=}N_v{=}256$, $N{=}24$). The inverse mapping from UV to the output image space is performed by a second STN with a bilinear kernel.
    \smallskip
    \item[] \textbf{Inpainting module}. The UV inpainting module has an autoencoder like architecture and consists of a number of part-specific encoders and decoders with gated embeddings... (extend).
    \smallskip
    \item[] \textbf{Blending module}. The blending module is similar in structure to the predictive module, but has 3 residual blocks and no downscaling or upscaling layers. As an input, it takes a concatenation of the source image and the target pose, as well as outputs of the predictive and the warping modules.\medskip
    }
    
    \textbf{Training parameters}. We train the network and its submodules with Adam optimizer with initial learning rate $2{\cdot}10^{-4}$ and $\beta_1{=}0.5$, $\beta_2{=}0.999$ (no weight decay). For speed, we pretrain the predictive module and the inpainting module separately and then train the blending network while finetuning the whole combined architecture end-to-end; DensePose network parameters remain fixed. In all experiments, the batch size is set to 8 and training proceeds for 40 epochs. The balancing weights $\lambda$ between different losses in the blending step (described in Sec.~\ref{losses}) are  set empirically to $\lambda_{\ell_1}{=}1$, $\lambda_{\text{p}}{=}0.5$, $\lambda_{\text{style}}{=}5{\cdot}10^{5}$, $\lambda_{\text{GAN}}{=}0.1$.

\setlength{\tabcolsep}{4pt}
\begin{table}[!t]
\begin{center}
\caption{Quantitative comparison 
with the state-of-the-art methods on the DeepFashion dataset~\cite{deepfashion} according to the Structural Similarity (SSIM)~\cite{SSIM}, Inception Score (IS) \cite{inceptionScore} and detection score (DS)~\cite{SiarohinSangineto2017} metrics. Our \textit{best structure} model corresponds to the $\ell_1$ loss, the \textit{highest realism} model corresponds to the style loss training (see the text and Table~\ref{tbl:losses}). Our \textit{balanced} model is trained using the full combination of losses.}
\label{table:deepfashion}
\begin{tabular}{lccc}
\hline\noalign{\smallskip}
\hspace*{10mm} Model & SSIM & IS & DS\\
\noalign{\smallskip}
\hline
\noalign{\smallskip}
Disentangled \cite{MaSunGeorgoulisCVPR2018}	     & 0.614   & 3.29 & --\\
VariGAN \cite{BoWuCheng2017}	     & 0.620   & 3.03 & --\\
G1+G2+D \cite{PoseGuided}                     & 0.762 & 3.09 &-- \\
DSC \cite{SiarohinSangineto2017}    & 0.761 & 3.39& 0.966 \\
\noalign{\smallskip}
\hline\noalign{\smallskip}
Ours (best structure)   & \textbf{0.796}  & 3.17 & 0.971\\
Ours (highest realism)  & 0.777  & \textbf{3.67} & 0.969\\
Ours (balanced)         & 0.785  & 3.61 & 0.971\\
\end{tabular}
\end{center}
\end{table}
\setlength{\tabcolsep}{1.4pt}

\begin{table}[!t]
\begin{center}
\caption{Results of the qualitative user study.}
\label{table:userstudy}
\begin{tabular}{lccccc}
\hline\noalign{\smallskip}
\multirow{2}{*}{\hspace*{10mm} Model} & \multicolumn{2}{c}{Realism} & \multicolumn{2}{c}{Anatomy} & Pose\\
& R2G & G2R & Real & Gen. & Gen. \\
\noalign{\smallskip}
\hline
\noalign{\smallskip}
 G1+G2+D [4]                     & 9.2 & 14.9  & -- & -- & -- \\
 DSC [5], experts    & 12.4 & 24.6 & -- & -- & -- \\ \noalign{\smallskip}\hline\noalign{\smallskip}
DSC [5], AMT    & 11.8 & \textbf{23.5} & 77.5 & 34.1 & 69.3\\
Our method, AMT & 13.1 & 17.2 & 75.6 & \textbf{40.7} & \textbf{78.2} \\\noalign{\smallskip}
\hline
\end{tabular}
\end{center}
\end{table}

\subsection{Evaluation metrics}
As of today, there exists no general criterion allowing for adequate evaluation of the generated image quality from the perspective of both structural fidelity and photorealism. We therefore adopt a number of separate structural and perceptual metrics widely used in the community and report our joint performance on them. 

\textbf{Structure}. The geometry of the generations is evaluated using the perception-correlated \textit{Structural Similarity} metric (SSIM)~\cite{SSIM}. 
We also exploit its multi-scale variant MS-SSIM~\cite{MSSSIM} to estimate the geometry of our predictions at a number of levels, from body structure to fine clothing textures.

\textbf{Image realism}. Following previous works, we provide the values of \textit{Inception scores (IS)}~\cite{inceptionScore}. However, as repeatedly  noted in the literature, this metric is of limited relevance to the problem of within-class object generation, and we do not wish to draw strong conclusions from it.
We have empirically observed its instability and high variance with respect to the perceived quality of generations and structural similarity. We also note that the ground truth images from the DeepFashion dataset have an average IS of 3.9, which indicates low degree of realism of this data according to the IS metric (for comparison, IS of CIFAR-10 is 11.2~\cite{inceptionScore} with best image generation methods achieving IS of 8.8~\cite{KarrasAilaLaine2017}).

In addition, we perform additional evaluation using \textit{detection scores (DS)}~\cite{SiarohinSangineto2017} reflecting the similarity of the generations to the~\textit{person} class. Detection scores correspond to the maximum of confidence of the PASCAL-trained SSD detector~\cite{SSD} in the person class taken over all detected bounding boxes.


\begin{figure}[!t]
\centering
\includegraphics[width=\linewidth]{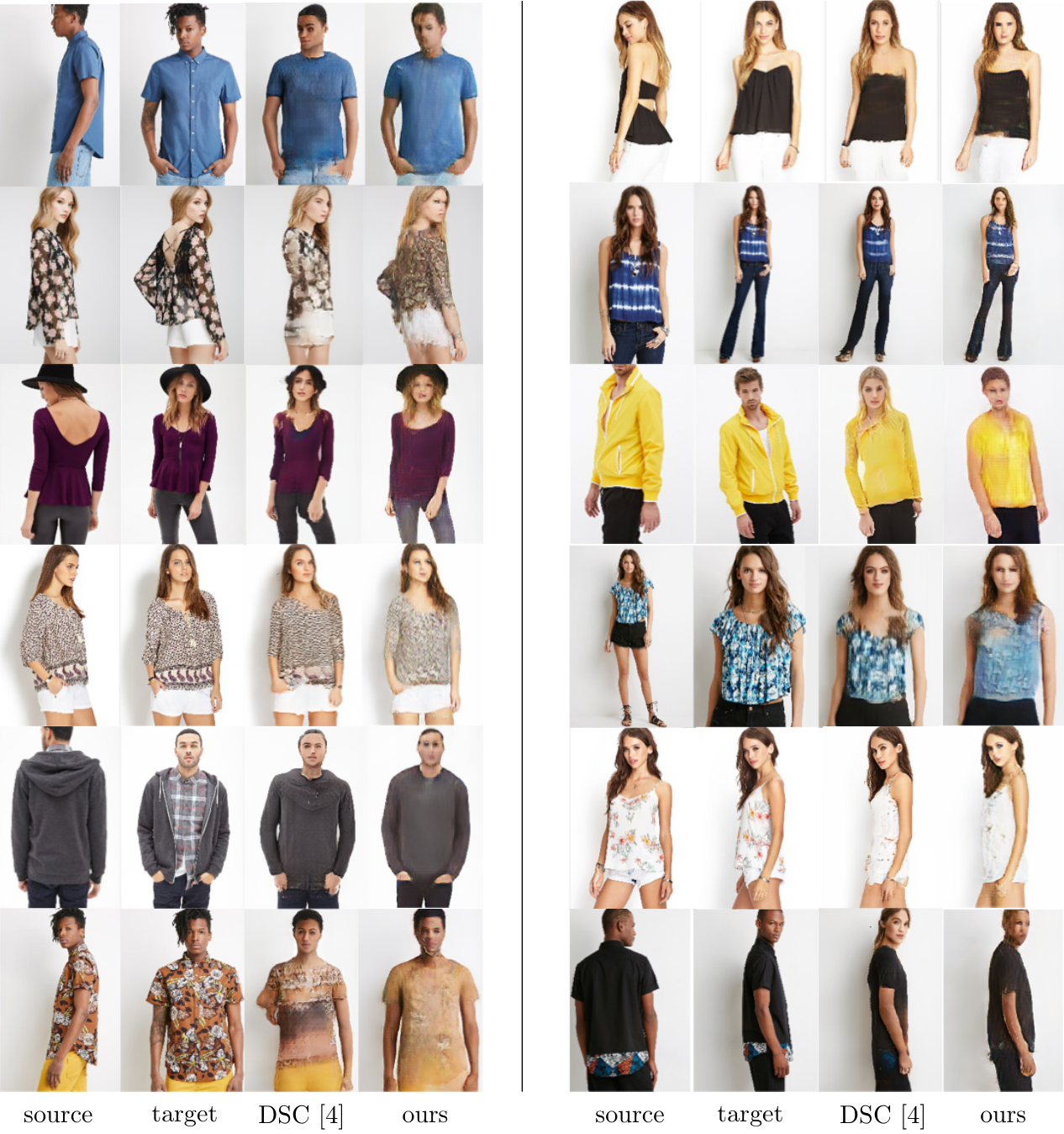}
\caption{Qualitative comparison with the state-of-the-art Deformable GAN (DSC) method of \cite{SiarohinSangineto2017}. Each group shows the input, the target image, DSC predictions~\cite{SiarohinSangineto2017}, predictions obtained with our full model. We observe that even though our cloth texture is occasionally not  as sharp, we better retain face, gender, and skin color information.}
\label{fig:results}
\end{figure}

\subsection{Comparison with the state-of-the-art}
We first compare the performance of our framework to a number of recent methods proposed for the task of \textit{keypoint guided} image generation or multi-view synthesis. Table~\ref{table:deepfashion} shows a significant advantage of our pipeline in terms of structural fidelity of obtained predictions. This holds for the whole range of tested network configurations and training setups (see Table~\ref{tbl:losses}). In terms of perceptional quality expressed through IS, the output generations of our models are of higher quality or directly comparable with the existing works. Some qualitative results of our method (corresponding to the \textit{balanced} model in Table~\ref{table:deepfashion}) and the best performing state-of-the-art approach~\cite{SiarohinSangineto2017} are shown in Fig.~\ref{fig:results}.

We have also performed a user study on Amazon Mechanical Turk, following the  protocol of~\cite{SiarohinSangineto2017}: we show 55 real and 55 generated images in a random order to 30 users for one second. As the experiment of~\cite{SiarohinSangineto2017} was done with the help of fellow researchers and not AMT users, we perform an additional evaluation of images generated by~\cite{SiarohinSangineto2017}
for consistency, using the official public implementation.
We perform three evaluations, shown in Table~\ref{table:userstudy}: \textit{Realism} asks users if the image is real or fake. 
\textit{Anatomy} asks if a real, or generated image is anatomically plausible. \textit{Pose} shows a pair of a target and a generated image and asks if they are in the same pose. The results (correctness, in \%) indicate that  generations of~\cite{SiarohinSangineto2017} have higher degree of perceived realism, but our generations show improved pose fidelity and higher probability of overall anatomical plausibility. 

\subsection{Effectiveness of different body representations}
In order to clearly measure the effectiveness of the DensePose-based conditioning, we first  compare to the performance of the `black box', predictive module when used in combination with more traditional body representations, such as background/foreground masks, body part segmentation maps or body landmarks.

\begin{figure}[t!]
\centering
\includegraphics[width=0.8\linewidth]{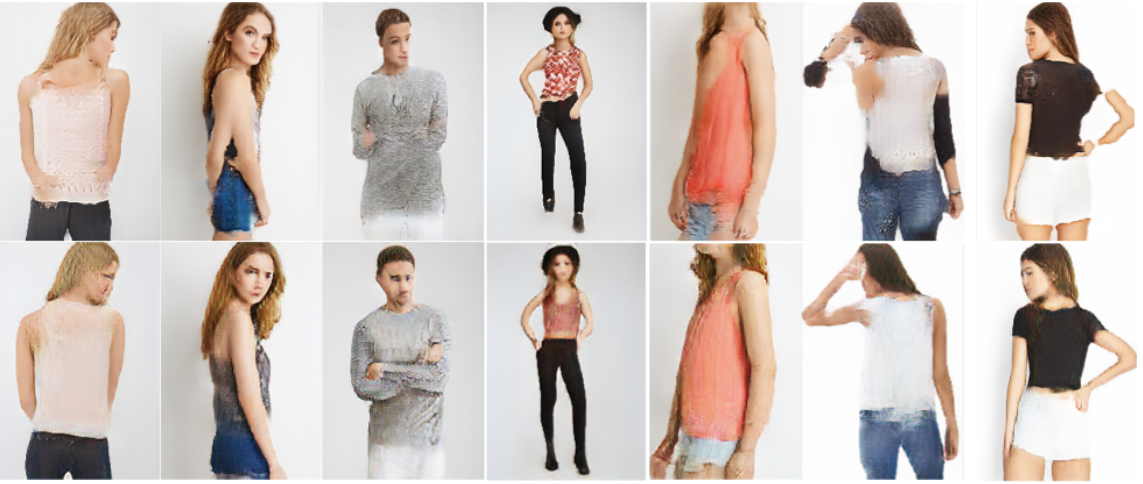}
\caption{Typical failures of keypoint-based pose transfer (top) in comparison with DensePose conditioning (bottom) indicate disappearance of limbs, discontinuities, collapse of 3D geometry of the body into a single plane and confusion in ordering in depth.}
\label{fig:geometry}
\end{figure}

\setlength{\tabcolsep}{4pt}
\begin{table}[!t]
\begin{center}
\caption{On effectiveness of different body representations as a ground for pose transfer. The DensePose representation results in the highest structural quality.}
\label{tbl:conditioning}
\begin{tabular}{lcccc}
\hline\noalign{\smallskip}
\hspace*{17mm}Model & SSIM & MS-SSIM & IS \\
\noalign{\smallskip}
\hline
\noalign{\smallskip}
Foreground mask            & 0.747 &  0.710 & 3.04           \\
Body part segmentation   & 0.776 & 0.791 & 3.22 \\
Body keypoints  & 0.762 & 0.774 & 3.09  \\
DensePose $\{I, U, V\}$  & \textbf{0.792} &  \textbf{0.821} & 3.09\\
DensePose $\{\text{one-hot}\,I, U, V\}$  & 0.782 &  0.799  & 3.32 \\
\end{tabular}
\end{center}
\end{table}
\setlength{\tabcolsep}{1.4pt}

As a segmentation map we take the index component of DensePose and 
use it to form 
a one-hot encoding of each pixel into a set of class specific binary masks.
Accordingly, as a background/foreground mask, we simply take all pixels with positive DensePose segmentation indices.
Finally, following \cite{SiarohinSangineto2017} we use~\cite{CMUdetector} to obtain body keypoints and one-hot encode them. 

In each case, we train the predictive module by concatenating the source image with a corresponding representation of the source and the target poses which results in 4 input planes for the mask, 27  for segmentation maps and 21 for the keypoints. 

The corresponding results shown in Table~\ref{tbl:conditioning} demonstrate a clear advantage of fine-grained dense conditioning over the sparse, keypoint-based,  or coarse, segmentation-based, representations.

Complementing these quantitative results, typical failure cases of keypoint-based frameworks are demonstrated in Figure~\ref{fig:geometry}. We observe that these shortcomings are largely fixed by switching  to the DensePose-based conditioning.

\setlength{\tabcolsep}{4pt}
\begin{table}[!t]
\begin{center}
\caption{Contribution of each of the functional blocks of the framework}
\label{tbl:blocks}
\begin{tabular}{lccc}
\hline\noalign{\smallskip}
\hspace*{30mm} Model & SSIM & MS-SSIM & IS   \\
\noalign{\smallskip}
\hline
\noalign{\smallskip}
predictive module only  & 0.792& 0.821 & 3.09 \\
predictive + blending (=self-refinement) & 0.793 & 0.821 & 3.10\\
predictive + warping + blending  & 0.789 & 0.814 & 3.12\\
predictive + warping + inpainting + blending (full)          & \textbf{0.796} & \textbf{0.823} & \textbf{3.17}
\end{tabular}
\end{center}
\end{table}
\setlength{\tabcolsep}{1.4pt}

\subsection{Ablation study on architectural choices}

Table~\ref{tbl:blocks} shows the contribution of each of the predictive module, warping module, and inpainting autoencoding blocks in the final model performance. For these experiments, we use only the reconstruction loss $\mathcal{L}_{\ell_1}$, factoring out fluctuations in the performance due to instabilities of GAN training. As expected, including the warping branch in the generation pipeline results in better performance, which is further improved by including the inpainting in the UV space. Qualitatively, exploiting the inpainted representation has two advantages over the direct warping of the partially observed texture from the source pose to the target pose: first, it serves as an additional prior for the fusion pipeline, and, second, it also prevents the blending network from generating clearly visible sharp artifacts that otherwise appear on the boarders of partially observed segments of textures.

\subsection{Ablation study on supervision objectives}
\begin{figure}[!t]
\centering
\includegraphics[width=.95\linewidth]{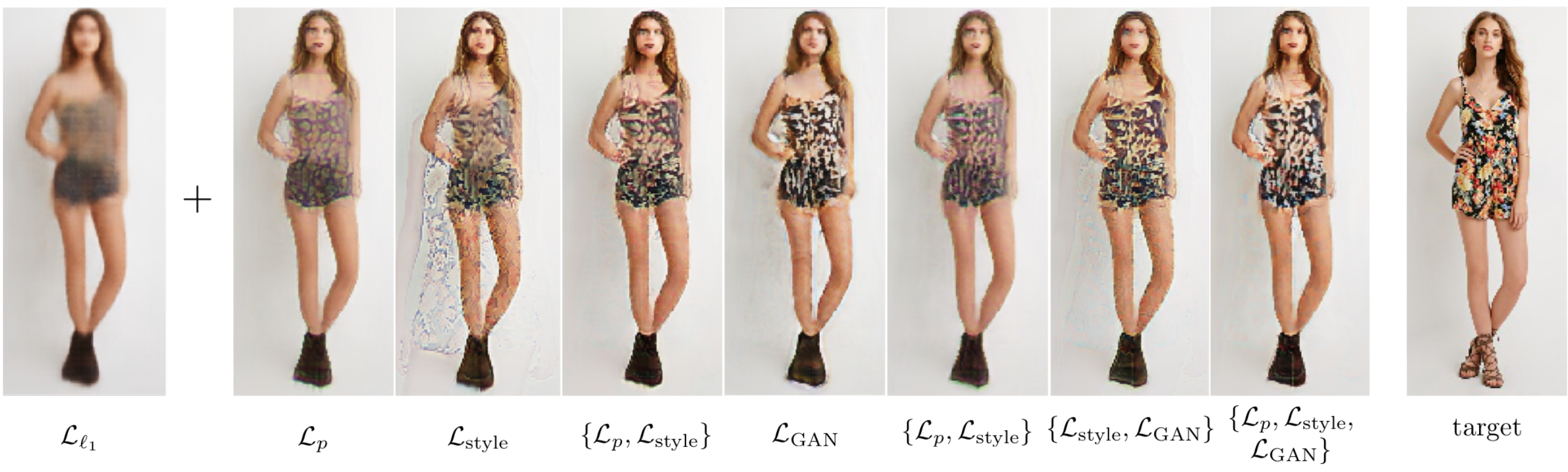}
\caption{Effects of training with different loss terms and their weighted combinations.}
\label{fig:losses}
\end{figure}

In Table~\ref{tbl:losses} we analyze the role of each of the considered terms in the composite loss function used at the final stage of the training, while providing indicative results in Fig.~\ref{fig:losses}.

\setlength{\tabcolsep}{4pt}
\begin{table}[t]
\begin{center}
\caption{Comparison of different loss terms used at the final stage of the training. The perceptual loss is best correlated with the structure, and the style loss with IS. The combined model  provides an optimal balance between the extreme solutions.}
\label{tbl:losses}
\begin{tabular}{lccc}
\hline\noalign{\smallskip}
\hspace*{15mm}Model & SSIM & MS-SSIM & IS   \\
\noalign{\smallskip}
\hline
\noalign{\smallskip}
$\{\mathcal{L}_{\ell_1}\}$ & \textbf{0.796} & \textbf{0.823} & 3.17    \\
$\{\mathcal{L}_{\ell_1},\,\mathcal{L}_{p}\}$ & 0.791 & 0.822 & 3.26    \\
$\{\mathcal{L}_{\ell_1},\,\mathcal{L}_{\text{style}}\}$ & 0.777 & 0.815 & \textbf{3.67}   \\
$\{\mathcal{L}_{\ell_1},\,\mathcal{L}_{p},\,\mathcal{L}_{\text{style}}\}$ & 0.784 & 0.820 & 3.41   \\
$\{\mathcal{L}_{\ell_1},\,\mathcal{L}_{\text{GAN}}\}$ & 0.771 & 0.807 & 3.39    \\
$\{\mathcal{L}_{\ell_1},\,\mathcal{L}_{p},\,\mathcal{L}_{\text{GAN}}\}$ & 0.789 & 0.820 & 3.33   \\
$\{\mathcal{L}_{\ell_1},\,\mathcal{L}_{\text{style}},\,\mathcal{L}_{\text{GAN}}\}$ & 0.787 & 0.820 & 3.32   \\
$\{\mathcal{L}_{\ell_1},\,\mathcal{L}_{p},\,\mathcal{L}_{\text{style}},\,\mathcal{L}_{\text{GAN}}\}$ & \textbf{0.785} & \textbf{0.807} & \textbf{3.61}   \\
\end{tabular}
\end{center}
\end{table}
\setlength{\tabcolsep}{1.4pt}

The perceptual loss $\mathcal{L}_p$ is most correlated with the image structure and least correlated with the perceived realism, probably due to introduced textural artifacts. 
At the same time, the style loss $\mathcal{L}_{\text{style}}$ produces sharp and correctly textured patterns while hallucinating edges over uniform regions.
Finally, adversarial training with the loss $\mathcal{L}_{\text{GAN}}$ tends to prioritize visual plausibility but often disregarding structural information in the input.
This justifies the use of all these complimentary supervision criteria in conjunction, as indicated in the last entry of Table~\ref{tbl:losses}.




\section{Conclusion}
In this work we have introduced a two-stream architecture for pose transfer that exploits the power of dense human pose estimation. We have shown that dense pose estimation is a clearly superior conditioning signal for data-driven human pose estimation, and also facilitates the formulation of the pose transfer problem in its natural, body-surface parameterization through inpainting. In future work we intend to further pursue the potential of this method for photorealistic image synthesis \cite{KarrasAilaLaine2017,ChenKoltun2017} as well as the treatment of more categories.

\bibliographystyle{splncs}
\bibliography{egbib}
\newpage

%
%
%
%

\section{Appendix A. Inpainting Autoencoder design}

 Table \ref{tab}  summarizes the architecture of our Inpainting Autoencoder's architecture. We note that we opted for a wide and shallow autoencoder architecture, because deeper networks seemed prone to overfitting, and intend to further explore alternative architectures and regularization methods, while also exploiting larger datasets.

\newcommand{\tensor}[2]{$ #1 \times #2 \times #2 $}
\newcommand{\kernel}[3]{$ #1 \times #2 \times 3 \times 3$ }
\newcommand{\myvector}[1]{$ 1  \times #1 $}
\newcommand{\linear}[2]{$ #1  \times #2 $}
\setlength{\tabcolsep}{4pt}
	\begin{table}
		\begin{center}
			\caption{Inpainting Autoencoder architecture -- every layer uses a linear operation, followed by a ReLU nonlinearity and Instance Normalization, omitted for simplicity. We indicate the size of the encoder's input tensor and the decoder's output tensor for symmetry. There are 24 separate encoder and decoder layers for the 24 parts, while the context layer fuses the per-part encodings into a common 256-dimensional vector.}
			\label{tab}
			\begin{tabular}{|l|c|c|}
				\multicolumn{3}{c}{Encoder (Input $T_i$, delivers $\mathcal{E}_i$)}\\\hline
				& Input tensor & Convolution Kernel  \\\hline
			Layer1	& \tensor{3}{256}  &  \kernel{3}{32}{2}\\\hline
			Layer2 	& \tensor{32}{128}  & \kernel{32}{32}{2}\\\hline
			Layer3 	& \tensor{32}{64}  & \kernel{32}{64}{2} \\\hline
			Layer4 	& \tensor{64}{32} & \kernel{64}{128}{2}\\\hline
				\multicolumn{3}{c}{Context network (Input $\{E_1,\ldots,E_{24}\}$, delivers $\mathcal{G}$)}\\\hline
& Input vector & Linear layer  \\\hline
					Layer 5 & \myvector{(24*128*16*16)} & \linear{(24*128*16*16)}{256}\\\hline
							Layer 6 & \myvector{256} & \linear{256}{256}\\\hline
							\multicolumn{3}{c}{Decoder (input $[\hat{E_i},G]$, delivers $\hat{T}_i$)}\\\hline
				& Output tensor &  Deconvolution kernel \\\hline
			Layer7 	& \tensor{64}{32} & \kernel{(128+256)}{64}{2}\\\hline
			Layer8 	& \tensor{32}{64} & \kernel{64}{32}{2}\\\hline
			Layer9 	& \tensor{32}{128} & \kernel{32}{32}{2}\\ \hline
			Layer10 	& \tensor{3}{256} & \kernel{32}{3}{2}\\\hline
			\end{tabular}
		\end{center}
	\end{table}
	\setlength{\tabcolsep}{1.4pt}
	\subsection{Training Loss  for Inpainting Autoencoder}

Injecting the global pose information,~$\mathcal{G}$ to the individual autoencoders through~$\mathcal{E}'_i$ makes it possible to transfer information from visible to unvisible parts. Still, this potential will not be exploited unless the supervision signal indicates that it is useful. From a single view of a person it is not possible to indicate e.g. that the front and the back of the torso typically have clothes of the same color. We therefore train our Inpainting Autoencoder with multiple views of the same person, which are reconstructed based on the single, input view. 

In particular, denoting by $I_i$, $P_i$ image $i$ and corresponding DensePose results, the input to our Inpainting Autoencoder is delivered by a Spatial Transformer Network (STN)~\cite{STN}:
\begin{equation}
T_i = \text{STN}(I_i,P_i)
\end{equation}
where the image $I_i$ is warped according to the deformation field $P_i$ and brought into correspondence with the surface-based UV system. 

In the DeepFashion dataset that we used for our experiments, each image is accompanied with a set of~$j{\in}\mathcal{N}(i)$  ``neighbor''  images of the same person wearing the same clothes but photographed with a different pose. These are warped to the common UV coordinate system as image $i$, i.e. $T_j = \text{STN}(I_j,P_j), \forall j \in \mathcal{N}(i)$ and used to form the reconstruction loss of the Inpainting Autencoder for the $i$-th image:
\begin{equation}
L_i =  \sum_{j \in {\mathcal{N}(i)}} \sum_{x \in \mathcal{S}(T_j)} |\hat{T}[x] - T_j[x] |, \quad \hat{T} = \text{IAE}(T_i)
\end{equation}
where $\hat{T}$ is the prediction of the Inpainting Autoencoder, $ \mathcal{S}(T_j)$ is the set of observable UV coordinates for image $j$, and we use the $\ell_1$ reconstruction loss on the observable part of each view. 

\section{Appendix B. Additional experiments on the DeepFashion dataset}

In order to further confirm the general merit of DensePose-based conditioning we performed an additional set of experiments, shown in Table~\ref{dpmerits}. In pariticular, the DensePose $\xrightarrow{}$ Keypoints row amounts to extracting DensePose on the input image, and keypoints on the target. We observe that while only using Keypoints on the target image, we still obtain better results, thanks to better capturing the pose of the input image.

\setlength{\tabcolsep}{4pt}
\begin{table}[!htb]
\begin{center}

\caption{\small Accuracy measures obtained by conditioning on different input/output pose estimators for the data-driven network; please see text for details.} 
\label{dpmerits}
\begin{tabular}{lcccc}
\hline
\noalign{\smallskip}
\hspace*{7mm}input $\xrightarrow{}$ target & SSIM & MS-SSIM & IS \\
\noalign{\smallskip}
\hline
\noalign{\smallskip}
Keypoints $\xrightarrow{}$ Keypoints  & 0.762 & 0.774 & 3.09  \\
\hline
Keypoints $\xrightarrow{}$ DensePose  & 0.779 & 0.796 & 3.10  \\
DensePose $\xrightarrow{}$ Keypoints  & 0.771 & 0.781 & 3.12  \\
DensePose $\xrightarrow{}$ DensePose  & 0.792 & 0.821 & 3.09  \\
\noalign{\smallskip}
\hline
\end{tabular}
\end{center}
\end{table}

\section{Appendix C. Experiments on the MVC dataset}

We provide results on the large scale MVC dataset~\cite{MVC} that consists of 161,260 images of high resolution crawled from several online shopping websites and showing front, back, left, and right views for each clothing item.

For our experiments, we downsample the images to the resolution of $256{\times}256$, so that they are commensurate with the rest of our experimental pipeline; the DensePose extractor is applied to their original $512{\times}512$ versions. We select 1000 clothing items for testing leaving the rest for training, which results into 694\,854 training pairs and 14\,924 test pairs. As in the case of the DeepFashion dataset, we include pairs of identical images in the training set.

We compare our obtained performance with state-of-the-art methods applied to this dataset in the setting of \textit{multi-view image synthesis} and show significant improvements both in terms of structural accuracy and realism (see Table~\ref{table:mvc}).
Some qualitative results of our method are shown in Fig.~\ref{fig:results}.

\setlength{\tabcolsep}{4pt}
\begin{table}[!t]
\begin{center}
\caption{Quantitative comparison of model performance with the state-of-the-art methods for multi-view image synthesis on the MVC dataset~\cite{MVC}.} 
\label{table:mvc}
\begin{tabular}{lcc|c}
\hline\noalign{\smallskip}
\hspace*{10mm} Model & SSIM & IS & MS-SSIM\\
\noalign{\smallskip}
\hline
\noalign{\smallskip}
cVAE \cite{SohnLeeYanNIPS2015}	     & 0.66   & 2.61 & --\\
cGAN \cite{MirzaOsindero2014}	     & 0.69   & 3.45 & -- \\
VariGAN \cite{BoWuCheng2017}	     & 0.70   & 3.69 & --\\
\noalign{\smallskip}
\hline\noalign{\smallskip}
Ours    & 0.85  & 3.74 & 0.89 \\
\hline\noalign{\smallskip}
\textit{Real data} & \textit{1.0} & \textit{5.47} & \textit{1.0} \\
\end{tabular}
\end{center}
\vspace{-0.6cm}
\end{table}
\setlength{\tabcolsep}{1.4pt}

\begin{figure}
\centering
\includegraphics[width=0.75\linewidth]{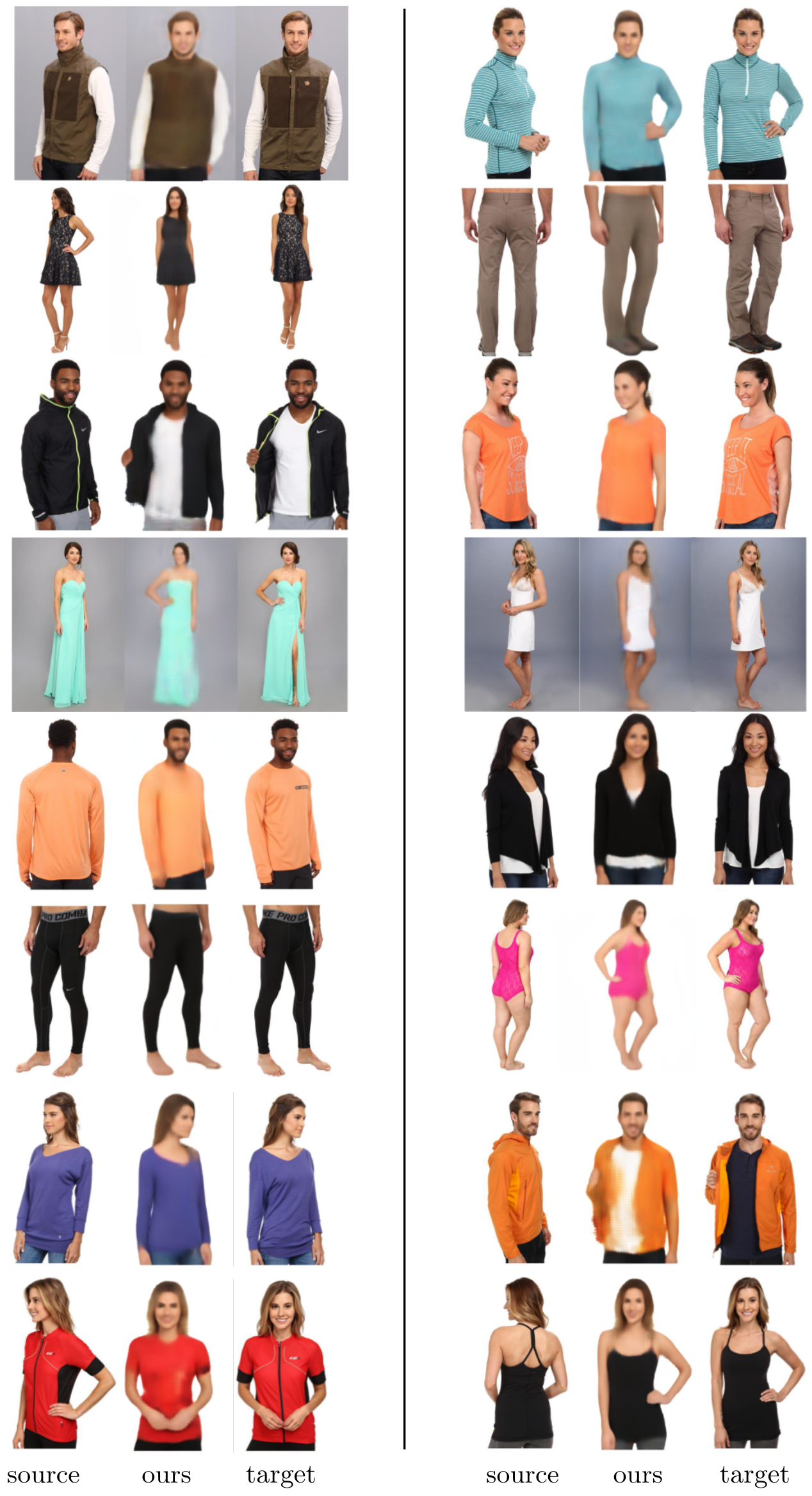}
\caption{Qualitative results on the MVC dataset (trained with l2 loss only).}
\label{fig:results}
\end{figure}

\clearpage

\end{document}